\documentclass[runningheads]{llncs}
\usepackage{graphicx}

\usepackage{csquotes}
\usepackage{paralist}
\usepackage{tikz}
    \usetikzlibrary{shapes, arrows.meta, positioning}
    \usetikzlibrary{shapes.geometric}
\usepackage{tabularx}
    \newcolumntype{e}{>{\hsize=.3\hsize\centering\arraybackslash}X}
    \newcolumntype{f}{>{\hsize=.5\hsize\arraybackslash}X}
    \newcolumntype{d}{>{\centering\arraybackslash}X}
\usepackage{threeparttable}
\usepackage{booktabs}
\usepackage{multirow}
\usepackage{smartdiagram}
\usepackage{lipsum}
\usepackage{setspace}
\usepackage{multicol}
\usepackage{color}
\usepackage{pdfpages}
\usepackage{pgfplots}
\usepackage{comment}
\usepackage{float}
\usepackage{enumitem}
\usepackage{xurl}
\usepackage{hyperref}
\usepackage{subcaption}
\usepackage{cancel}
\MakeOuterQuote{"}

\begin{document}
\title{The Use of Artificial Intelligence in Military \\ Intelligence: An Experimental Investigation of \\ Added Value in the Analysis Process}
\titlerunning{The Use of Artificial Intelligence in Military Intelligence}
\author{
Christian Nitzl \inst{1,2} 
\and Achim Cyran \inst{1} 
\and Sascha Krstanovic \inst{3} 
\and Uwe M. Borghoff \inst{4}
}
\authorrunning{C. Nitzl, A. Cyran, S. Krstanovic \& U. M. Borghoff}
%
\institute{Center for Intelligence and Security Studies (CISS) \\
University of the Bundeswehr Munich, Neubiberg, Germany \\
https://www.unibw.de/ciss 
\and
Department of Economics and Management \\ University of the Bundeswehr Munich,
Neubiberg, Germany \\ christian.nitzl@unibw.de
\and 
Aleph Alpha, Heidelberg, Germany \\ https://aleph-alpha.com
\and
Institute for Software Technology \\
University of the Bundeswehr Munich, Neubiberg, Germany \\
Corresponding author: uwe.borghoff@unibw.de
}
\maketitle              

\begin{abstract}
It is beyond dispute that the potential benefits of artificial intelligence (AI) in military intelligence are considerable. Nevertheless, it remains uncertain precisely how AI can enhance the analysis of military data. The aim of this study is to address this issue. To this end, the AI demonstrator deepCOM was developed in collaboration with the start-up Aleph Alpha. 

The AI functions include text search, automatic text summarization and Named Entity Recognition (NER). These are evaluated for their added value in military analysis. It is demonstrated that under time pressure, the utilization of AI functions results in assessments clearly superior to that of the control group. Nevertheless, despite the demonstrably superior analysis outcome in the experimental group, no increase in confidence in the accuracy of their own analyses was observed. Finally, the paper identifies the limitations of employing AI in military intelligence, particularly in the context of analyzing ambiguous and contradictory information.
\keywords{Military Intelligence \and Artificial Intelligence \and Open Source Intelligence \and Analysis Process \and Experiment.}
\end{abstract}

\section{Introduction}
The sheer volume of data that can be observed today makes it clear that military intelligence requires the use of artificial intelligence (AI) \cite{gartin2019future}. However, the benefits of using AI and at what point in the military analysis process is still an open question \cite{vogel2021impact}. The primary role of military intelligence is to gather and analyze information to support military leaders in making informed decisions. From an academic standpoint, military intelligence represents a transdisciplinary field of research that draws upon a multitude of disciplines, including political science, economics, sociology, and psychology, among others \cite{albrecht2022transdisciplinary}. 

Military intelligence is thus concerned with the collection and analysis of information to provide a comprehensive picture of the situation. This may entail the collection of data on the armed forces and the examination of the plans and operations of other nations, as well as the gathering of information on developments affecting a nation's security \cite{sadiku2021military}.

What is certain is that the use of innovative approaches and methods, such as artificial intelligence (AI), must be ensured when analyzing militarily relevant developments abroad. New developments in AI and its integration into analysis and research software promise a wide range of support options to enhance analysts' ability to make judgments \cite{cho2020priority}. 

It is expected that the use of AI technologies will reduce the burden on analysts, allowing them to focus on the core content of analyzing, assessing and presenting the military intelligence situation \cite{hare2016future}. 

It should be emphasized that the analyst should not be replaced by AI systems, but rather assisted. In particular, it must be ensured that analysts are always able to understand the information on which they are making an assessment \cite{blanchard2023ethics}.

As part of this study, a proprietary AI demonstrator was developed by the start-up company Aleph Alpha. The capabilities of this program called deepCOM are based on a Large Language Model (LLM). It should be emphasized that deepCOM is not a working product, but a demonstrator. The core functionality of deepCOM is semantic search. This allows the user to ask direct questions which are answered by the system, indicating the sources used. In addition, deepCOM can automatically summarize each report in the database, allowing the analyst to identify relevant sources from a summary of a few sentences. 

An additional Named Entity Recognition (NER) implemented in the system labels all reports fully automatically: if present in the text, tags are derived from mentions of time, places, organizations and people, which are highlighted for the user both when identifying relevant sources and when reading \cite{devlin2018bert}.

The goal of this study is to demonstrate the added value of using AI in the military analysis process. While previous studies have focused primarily on the use of AI in data collection \cite{horlings2023dealing}, this study focuses on the support AI provides to human analysis and assessment. The use of new technologies for their own sake is not desirable if it does not lead to direct added value for the analyst and his or her analytical performance. 

Conceptual considerations alone are not sufficient to assess value. In order to be able to make empirically validated statements, an experiment was conducted in this study. To the best of our knowledge, this is the first study to empirically analyze the added value of AI in the context of intelligence.

\newpage

This research question will be addressed using the following approach.
Section~\ref{sec2} provides an overview of the military analysis process based on the intelligence cycle. Section~\ref{sec3} then presents the AI functions that were investigated and how they support military analysts. Section~\ref{sec4} explains the experimental design, while Section~\ref{sec5} presents the resulting findings. Section~\ref{sec6} discusses the results of the experiment. Finally, Section~\ref{sec7} offers concluding remarks.

\section{The intelligence cycle as the starting point of the military analysis process}\label{sec2}
In order to assess the support provided by AI systems in the military analysis process, it is first necessary to clarify how the intelligence process works in general \cite{horlings2023dealing}. The starting point is the so-called intelligence cycle, which describes the ideal process from the decision maker's request for information to the intelligence product \cite{lowenthal2022intelligence}. It should be noted that the process is not a linear sequence of individual steps, but includes feedback loops \cite{hulnick2006s}. 

\begin{figure}[ht]
\centering\includegraphics[width=0.8\textwidth]{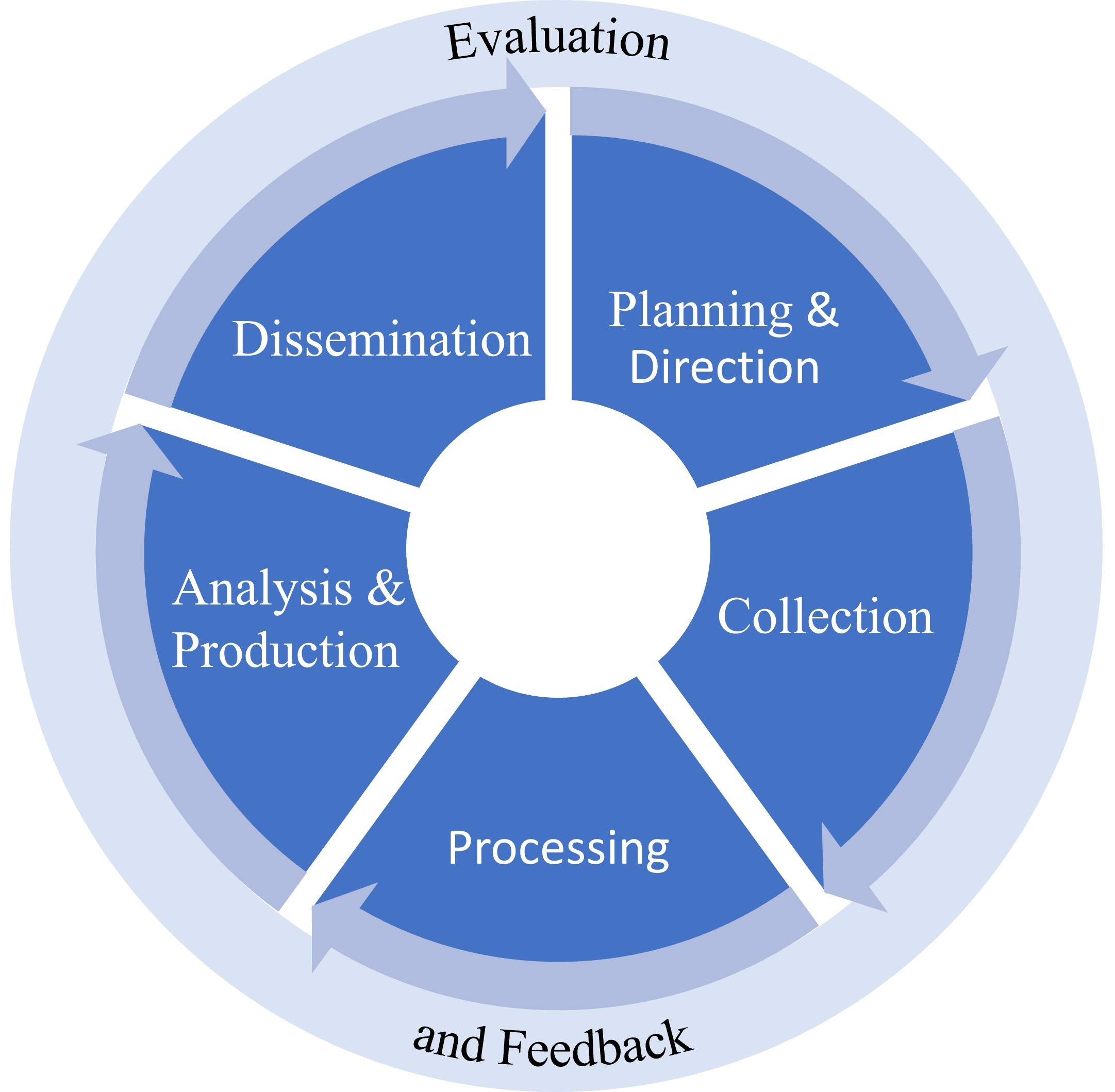}
\caption{Intelligence Cycle \cite{werro2024role}.} \label{Abb1}
\end{figure}

Figure~\ref{Abb1} illustrates the typical intelligence cycle.

The process begins with the \textit{Planning \& Direction} phase. In this phase, a client or customer, in our cases a military decision maker, formulates a need for intelligence. This need is usually expressed in terms of questions that the customer believes must be answered in order to be able to make an informed decision. This defines an intelligence problem \cite{clark2019intelligence,phythian2013understanding}.

Once the mandate is given, the second phase of the cycle, \textit{Collection}, begins. This involves gathering information needed to produce the finished intelligence product \cite{phythian2013understanding}. Today, collection can be based on a variety of different intelligence disciplines: Human Intelligence (HUMINT), Imagery Intelligence (IMINT), Signals Intelligence (SIGINT), and Open Source Intelligence (OSINT) \cite{clark2013intelligence,Nato2016}.

The Collection phase is followed by the \textit{Processing} phase, in which the collected information is processed \cite{phythian2013understanding}. This includes translating foreign language texts, decoding, and organizing the information from human sources into a standardized reporting format \cite{clark2013intelligence}. The main challenge in processing is that there is often more data from different sources than can be processed in a reasonable amount of time \cite{johnson1986making}.

The intelligence product is created in the \textit{Analysis \& Production} phase. This is done by integrating, evaluating, and analyzing all available information into an overall picture, taking into account the knowledge already available \cite{phythian2013understanding}. The analyst faces the challenge that the available information may be incomplete and contradictory. The goal of this phase is to obtain an assessment of ambiguous events and possible future events, thus providing the customer with a basis for an informed decision, for example, by recommending a course of action \cite{clark2019intelligence}.

Finally, in the \textit{Dissemination} phase, the intelligence product is distributed to the client. This can take the form of a written report or a verbal briefing \cite{clark2013intelligence}. The decisions made by the customer may directly lead to further intelligence requirements or at least influence the content requirements for future finished intelligence, so the circle from Dissemination to Planning \& Direction closes at this point \cite{phythian2013understanding}.

\section{AI capabilities in the deepCOM demonstrator in support of military intelligence}\label{sec3}
The deepCOM demonstrator is an analysis tool with integrated AI capabilities designed to support the work of military analysts. The AI functions experimentally analyzed are described below. 
Two of the three AI functions tested within deepCOM, namely AI search and automated summarization, are based on a Large Language Model (LLM). The third AI function tested is Named Entity Recognition. Although the intelligence community in Germany works in English due to international structures such as NATO, the United Nations and the EU, its own products are created in German. Accordingly, deepCOM's user interface and output are in German.

\subsection{Artificial intelligence search in text databases}
Standard searches in text databases are based on the frequency of occurrence of words. Accordingly, texts that contain more of the search terms will appear higher up on the list. In this method, also known as Bag of Words (BOW), the sheer frequency of the individual words determines a good search hit, not their relationship to each other \cite{qader2019overview}. The BOW search typically starts with a question in the user's mind, which the user must break down into several keywords \cite{BohneRB11}, rather than typing the entire question into the search box. This type of search is usually the only way for the military analyst to search text, as the text databases need to be in a secure environment.

Such an approach is inefficient for several reasons: Firstly, information is lost during the interaction between the user and the query due to the forced reduction to keywords. Even if the search still works despite the omission of prepositions, cases, numbers and conjugations, the information it contains can help to produce better search results. Furthermore, the search process is not intuitive. When a question is already formulated, it is more straightforward to enter it into the search box without making any modifications. This is a standard practice for all major search engines on the Internet. Finally, BOW's search is based exclusively on the frequency of words in texts. However, this approach may result in the retrieval of irrelevant documents that contain the keywords in question, despite the documents' lack of relevance to the user's actual query.

The problems of BOW search can be solved by AI search. AI search is able to process a question as a whole. This results in less loss of information. In deepCOM, this is achieved by first displaying the answer of the AI search in the output, which is either formulated by the system itself or taken directly from a text, depending on the complexity of the question. Since the way in which the LLM behind the AI search generates the answer is not always comprehensible or correct, the full-text passages from which the information originates are also referenced. As can be seen in Figure~\ref{Abb2}, the answer given by the LLM-based AI search is independent of the exact wording of the questions. In addition, the AI search can deal with different spellings of identical entities (in the example shown, the different transliteration of an Arabic proper name).

\begin{figure}[ht]
\centering\begin{threeparttable}
        \begin{tabularx}{.87\textwidth}{X p{0.2cm}}
        \toprule
        \multicolumn{2}{l}{\textbf{Answer: with Tomahawk missiles}} \\
        \multicolumn{2}{l}{Based on:} \\
        \midrule
        Mehr als 72 Stunden vergingen zwischen den sengenden Bildern eines Che\-mie\-waffenangriffs aus Syrien und der Reaktion des ame\-rikanischen Mi\-li\-tärs. Präsident Trump kündigte am Donners\-tag\-abend an, dass Dutzende von {\em Tomahawk-Raketen} auf den Flugplatz Al Shayrat abgefeuert worden seien, von dem aus am Diens\-tag der Angriff auf die Stadt Khan Sheikhoun gestartet worden war. & \\
        \bottomrule
        \end{tabularx}
        \vspace{0.5cm}
        \begin{tabularx}{.87\textwidth}{X p{0.2cm}}
        \toprule
        \multicolumn{2}{l}{\textbf{Answer: with Tomahawk cruise missiles}} \\
        \multicolumn{2}{l}{Based on:} \\
        \midrule
        Es klingt so einfach, aber so oft ist unsere politische Rhetorik abstrakt und lässt die menschlichen Opfer des Krieges außer Acht. Als Präsident Donald Trump am Donnerstag, den 6. April, den Abschuss von 59 {\em Tomahawk-Marschflugkörpern} auf den Flugplatz Al Shayrat in Syrien ankündigte, sagte er, er habe einen gezielten Militärschlag gegen den Flugplatz in Syrien angeordnet, von dem der Chemiewaffenangriff ausging. & \\
        \bottomrule
        \end{tabularx}
    \end{threeparttable}
\caption{Sample AI search answers to the questions 'How was the US air strike on Ash Sha`irat carried out?' and 'How did the US attack Al Shayrat airfield?'.} \label{Abb2}
\end{figure}

\subsection{Named Entity Recognition}
Named Entity Recognition (NER) refers to the extraction of entities from unstructured text and their classification into predefined categories \cite{lample2016neural}. The NER implemented in deepCOM is based on a German retraining of the Bidirectional Encoder Representations from Transformers models originally published by Google \cite{devlin2018bert,yadav2019survey}. It automatically identifies time, place, organization, and person entities, even without optimization for specific text corpora. Since entities in texts usually occur in inflected form, lemmatization is also required to convert them to their base form and thus make them comparable (e.g., Mittelmeers, Mittelmeere 
$\rightarrow$ Mittelmeer). Both works largely correctly, although rare entities are sometimes incorrectly categorized or incorrectly converted to an uninflected base form. A detailed description of the use of NER in military intelligence can be found in \cite{NitzlCKB24}.

In military intelligence, source code is still sometimes labeled manually, which is time consuming. NER can help the analyst to get a first impression of the content of a source. They can then more quickly decide whether a source is of interest. The color coding also speeds up the search for relevant information when reading full text. Finally, the automatically extracted entities can be displayed as locations on a world map in deepCOM. This allows the analyst to quickly locate events and integrate them into a situational picture. A heat map can also be used to identify clusters of locations and associated events. 

\begin{figure}[h]
\centering\includegraphics[width=.95\textwidth]{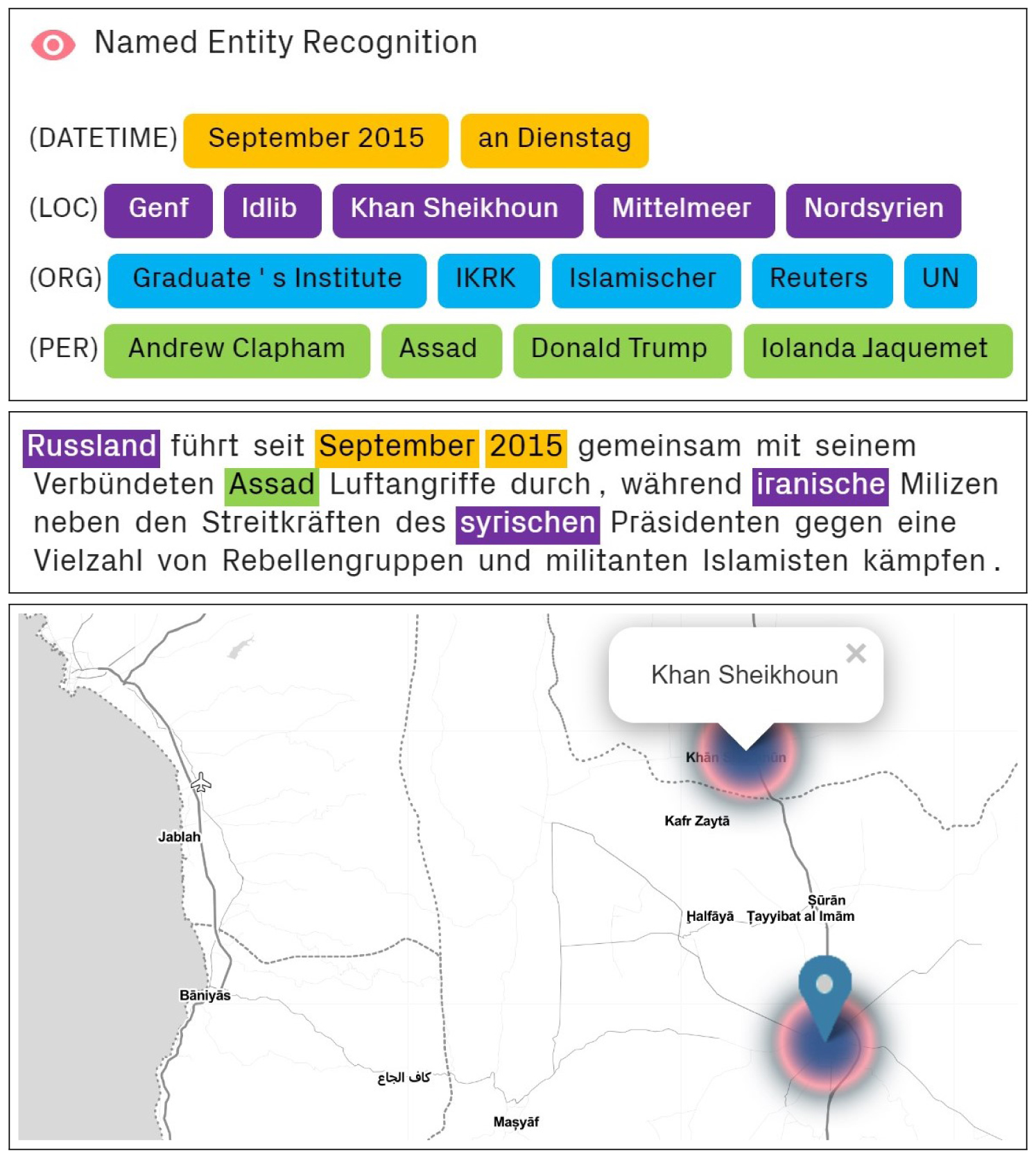}
\caption{Top image: NER automatically extracts time, place, organization, and person names from the text. Middle image: Color coding of recognized entities in the text. Bottom image: Display of recognized locations on a map.} \label{Abb3}
\end{figure}

Figure~\ref{Abb3} shows an example of the use of NER for the above explanations.

\subsection{Automated text summaries}
Irrespective of the possible full-text presentation, the references in deepCOM are also reduced to about one-third to one-half of their original length by summarizing each paragraph into one sentence. Automated summarization thus serves a similar purpose to NER. Both allow a quick assessment of the importance of a text for an analysis. The conflict between the length of the text and the depth of detail that summarization provides must be decided on a situation-by-situation basis. 

\newpage

Automatic summarization is enabled by the LLM in the deepCOM back-end. The neural network algorithm allows to merge paragraphs or omit parts of sentences in such a way that the summary is a coherent image of the original text. 

The implemented automatic summarization primarily uses the omission of individual sentence parts. Further tests with different settings for text summarization have shown that the summary generally works very well, but that too much world knowledge is added from the LLM training data.

\begin{figure}
\fbox{\parbox{.98\textwidth}{\textbf{Original:} \\ 
    Bei einem Luftangriff auf die von Rebellen kontrollierte Kleinstadt Khan Sheikhoun im Nordwesten Syriens ist am Dienstagmorgen ge\-gen 7 Uhr (6 Uhr in Frankreich) ein bislang unbekanntes Gas freigesetzt worden.
    \begin{center}
        \begin{tikzpicture}
            \draw[-{Triangle[width=10pt,length=15pt]}, line width=7pt](0,0) -- (0, -1);
        \end{tikzpicture}
    \end{center}
    \vspace{-0.5cm}
    \textbf{Summary:} \\ 
    Ein Luftangriff auf Khan Sheikhoun hatte ein nicht identifiziertes Gas freigesetzt. 
    \strut}} 
\caption{Example of an automated text summary: The text 'On Tuesday morning at around 7am (6am in France), an air strike on the small rebel-held town of Khan Cheikhoun in northwestern Syria released an as yet unidentified gas' is summarized as 'An air strike on Khan Cheikhoun released an unidentified gas'.} \label{Abb4}
\end{figure}
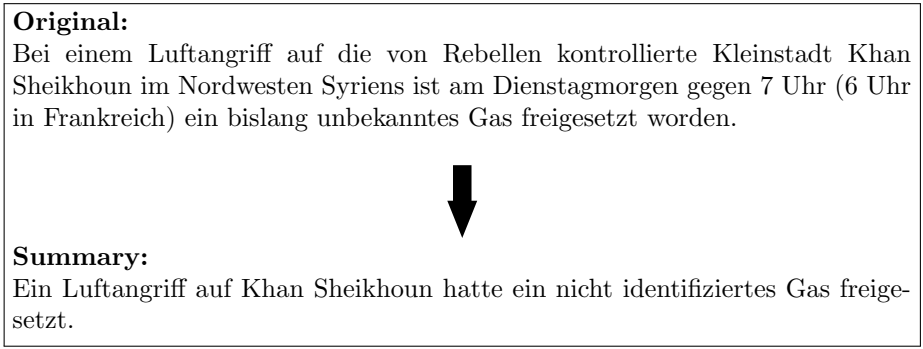

Figure~\ref{Abb4} shows an example of automated summarization.

\section{Description of the experimental study}\label{sec4}
\subsection{Military analysis scenario}
The starting point of the experimental study is a realistic scenario from military intelligence analysis. A total of 50 source texts was collected from publicly accessible news portals on the Internet. The sources were stored in deepCOM and served as a text database. The sources were selected to provide a comprehensive picture, including articles from both national and international news sites. These included news portals publishing in French, Russian, Arabic and Persian. 

An overview of the sources used can be found in Annex~\ref{annex1}.

\begin{table}[ht]
\caption{Confidence level and probability statements for assessments.\bigskip} \label{Tab1}
\centering\begin{threeparttable}
        \label{tab:Confidence_Probability}
        \begin{tabularx}{.9\textwidth}{p{3.5cm}X}
        \toprule
        \multicolumn{2}{l}{\textbf{Confidence Level}} \\
        \midrule
        High & Good quality of information, evidence from \newline multiple collection capabilities, possible to make \newline a clear judgement. \\
        Moderate & Evidence is open to a number of interpretations, \newline or is credible and plausible but lacks correlation. \\
        Low & Fragmentary information, or from collection \newline capabilities of dubious reliability. \\
        \bottomrule
        \end{tabularx}
        \vspace{0.5cm}
        \begin{tabularx}{.9\textwidth}{p{3.5cm}X}
        \toprule
        \multicolumn{2}{l}{\textbf{Probability statements for assessments}} \\
        \multicolumn{2}{l}{\textbf{(numerical and verbal)}} \\
        \midrule
        More than 90\% & Highly likely \\
        60\% - 90\% & Likely \\
        40\% - 60\% & Even chance \\
        10\% - 40\% & Unlikely \\
        Less than 10\% & Highly unlikely \\
        \bottomrule
        \end{tabularx}
    \end{threeparttable}
\end{table}

The news texts refer to the poison gas attack in Khan Shaykhun in the Idlib governorate in northwestern Syria. On April 4, 2017, at least 86 people were killed and several hundred injured by sarin gas. The release of the poison gas is uncontested, but explanations of how it happened vary widely: According to the US, UK, French and German accounts, the gas was deliberately dropped by an air strike by the Assad government's Syrian air force. The Idlib governorate and the town of Khan Shaykhun were considered a stronghold of the Syrian government's opponents at the time of the incident. However, according to Syrian, Russian, and Iranian accounts, the sarin was released because the Syrian Air Force had bombed an insurgent poison gas storage facility or factory with conventional weapons. In response to the poison gas attack, the US, under President Trump, launched cruise missile strikes on the Syrian military airfield of Al Shayrat, from which the attack on Khan Shaykhun is believed to have originated.
The central task in the analysis process is to identify and select the relevant sources that provide the necessary information for a correct assessment of the situation. In order to meet this challenge in the experiment, the 50 source texts were selected in such a way that about one third could be used directly for the analysis, another third dealt with the poison gas attack in Khan Shaykhun only in passing (e.g., mention in stock market news), and the last third had no reference to the incident to be analyzed. What all texts have in common, however, is that they contain the keywords 'Syria' and 'poison gas'. The last third of the texts is deliberately used as a distraction in both the BOW and AI searches in order to divert attention from the relevant topic.

The analysis task begins four days after the poison gas attack in Khan Shaykhun on April 7, 2017. A military leader needs detailed information about the poison gas attack in Syria and its aftermath in order to make a decision on how to proceed, and therefore wants to be briefed in writing about the developments so far. In order to specify his informational needs, he has defined several questions that are relevant to him. 
The participants should answer these questions in their own words in the first part of the analysis task. In order to achieve the best possible comparability between the experimental and the control groups, these questions must be answered in a few key points. In addition, at least one source must be cited for each answer. On the one hand, the sources mentioned by the participants reflect the reality in practice. On the other hand, it can be excluded that the participants arrive at the correct answers by guessing or through prior knowledge of the poison gas attack in Khan Shaykhun.

In the second part of the analysis task, participants were presented with theses the probability of occurrence of which had to be judged, see also Table~\ref{Tab1}. Participants were asked how likely they thought it was that the event described in the thesis had occurred or would occur. Answers could be given on a five-point scale from highly likely to highly unlikely. In addition, respondents were asked to indicate how confident they were in their own assessment of the sources (confidence). To do this, respondents had to answer a question about how confident they were in their own assessment given the information available. The response options were a three-point scale from high to medium to low. This procedure and the questioning of the scales follow the specifications of the NATO Allied Joint Doctrine for Intelligence Procedures (NATO, 2016).

Figure~\ref{Abb5} shows examples of items from both parts of the analysis task. 
The complete list of all items can be found in Annex~\ref{annex2}.

\begin{figure}[h]
\centering\fbox{\parbox{.98\textwidth}{\textbf{Part 1} (Please answer in your own words) \begin{compactenum}[(a)]
    \item Who controlled the region of the poison gas attack at the time?
    \item Who does Germany hold responsible for the poison gas attack?
    \item How did Russia react to the US air strike?
    \end{compactenum}}}
    \fbox{\parbox{.98\textwidth}{\textbf{Part 2} (Indication of probability and confidence) \begin{compactenum}[(a)]
    \item The chemical gas released is exclusively sarin.
    \item  Syria will demonstrate its independence and freedom of action to the international community through a (possible further) air strike with chemical warfare agents.
    \item Under international pressure, Syria will admit (co-)responsibility for the attack in Khan Shaykhun. 
    \end{compactenum}}}
\caption{Examples of tasks for the two parts of the analysis task.} \label{Abb5}
\end{figure}

\subsection{Design of the experiment}
Participants in the study were randomly assigned to either the experimental or control group. Both groups used the same browser interface, except that the AI functions were disabled in the control group. The AI search was replaced by a BOW search, and the automatic summarization was replaced by a display of the first words of a paragraph. The completion time for the analysis task was set to 30 minutes. An expert survey was conducted as a basis for evaluating the analysis performance of the participants. Experienced military analysts worked on the same analysis task as the participants in the experiment. They worked under the same conditions as the control group. However, the experts were not given a time limit to complete the analysis scenario. 

Any information that could be used to personally identify individual participants was deleted after data collection was completed.

Vouchers were raffled off to encourage participation. Of the original 30 participants, one person had to be excluded for technical reasons. Therefore, the experimental group consisted of 14 participants and the control group consisted of 15 participants. All participants are active duty soldiers. They range in age from 20 to 33 years old (M=26.6; SD=4.3) and 76\% are male. In addition, seven military intelligence experts completed the analysis task. They were between 31 and 57 years old (M=41.1; SD=8.0) and all male. On average, they had been soldiers for 21.5 years (SD=8.4). On average, the experts took 3 hours and 49 minutes to complete the analysis task. 

\section{Analysis of the experimental results}\label{sec5}
In a first step, the average score per item for the first part of the analysis task was calculated separately for the experimental and control groups. This was used to determine how close the participants in both groups were to the expert judgment. As the 21 items in total had different scores to be achieved due to their varying complexity, these are presented in relative terms. 

\begin{figure}[ht]
\begin{tikzpicture}
        \pgfplotsset{scale only axis}
        \begin{axis}[
            xmin=1, xmax=21,
            ymin=0, ymax=110,
            xlabel=Item,
            ylabel={Percentage of points achieved},
            width={.87\textwidth},
            height={0.5\textwidth},
            grid=major,
            xtick={1,2,...,21},
            every axis plot/.append style={thick}
            ]
            \addplot[smooth,mark=*]
                coordinates{
                (1,60.71)(2,82.14)(3,71.43)(4,46.43)(5,50.00)(6,53.57)(7,85.71)(8,100.00)(9,100.00)(10,71.43)(11,78.57)(12,85.71)(13,50.00)(14,35.71)(15,50.00)(16,50.00)(17,50.00)(18,40.48)(19,17.86)(20,28.57)(21,28.57)
                };
            \addplot[smooth,mark=square,blue]
                coordinates{
                (1,53.33)(2,40.00)(3,60.00)(4,33.33)(5,46.67)(6,53.33)(7,33.33)(8,60.00)(9,80.00)(10,40.00)(11,46.67)(12,46.67)(13,46.67)(14,20.00)(15,30.00)(16,24.44)(17,13.33)(18,17.78)(19,26.67)(20,6.67)(21,10.00)
                }; 
            \addlegendimage{/pgfplots/refstyle=plot_Drang}\addlegendentry{\small Experimental group},
            \addlegendimage{/pgfplots/refstyle=plot_Bewegung}\addlegendentry{\small Control group},
        \end{axis}
        
    \end{tikzpicture}
\caption{Percentage mean value per item separately for experimental and control groups in the first part of the analysis task.} \label{Abb6}
\end{figure}
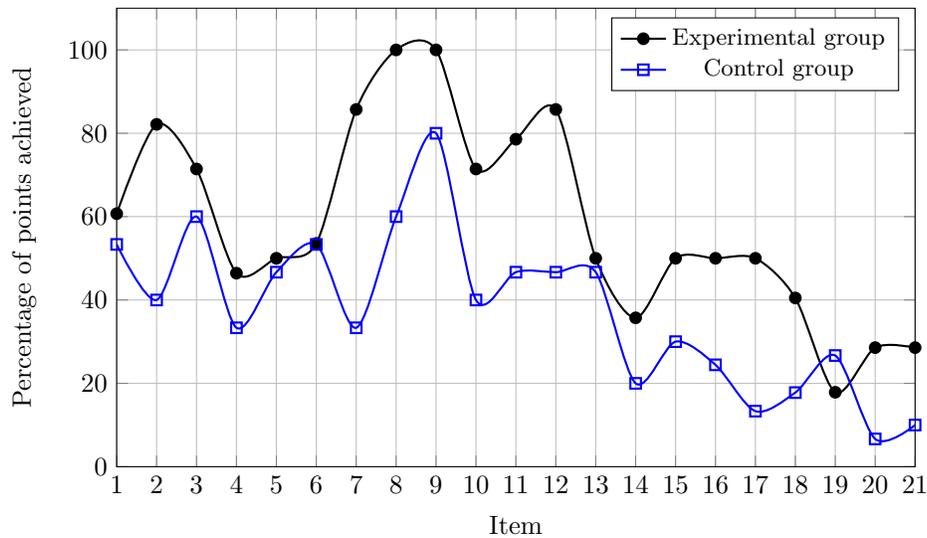

Figure~\ref{Abb6} shows the average percentages achieved per item. Since a score closer to 100\% means a higher average agreement with the expert judgment, a higher score is considered to be better in this figure.

The experimental group always performed better than the control group, except for Item 19. Furthermore, the items were of varying difficulty, as the curves are similar for both groups. For example, the solutions for Items 4 to 6 turned out to be significantly poorer than for Items 8 and 9, regardless of whether the AI functions were available to the subjects or not. Finally, it is noticeable that both groups were able to solve fewer and fewer items towards the end of the first part. This is due to the 25-minute time limit for the first part of the analysis task.

The next step was to test the related items from the first part of the analysis task as blocks. Task 6, which had an above-average number of six items and dealt with both the US air strike and the reaction of several nations to it, was split into two subtasks. Each of the now seven tasks were statistically tested for mean differences in independent samples. 

Table~\ref{Tab2} shows the means (M) and standard deviations (SD) of the experimental and control groups. The $\chi^2$ and p-values (p) are also shown in the table.

\begin{table}[t]
\caption{Participants' scores on the first part of the analysis task, separated into total performance and performance per task.\bigskip} \label{Tab2}

    \renewcommand{\arraystretch}{1.2}
    \setlength{\tabcolsep}{0.25cm}
    \begin{threeparttable}
        \begin{tabularx}{\textwidth}{X rrrrr l }
        \toprule
        & \multicolumn{2}{c}{\textbf{EG}} & \multicolumn{2}{c}{\textbf{KG}} & &  \\
        \cmidrule(rl){2-3} \cmidrule(lr){4-5}
        & \multicolumn{1}{c}{\textbf{\textit{M}}} & \multicolumn{1}{c}{\textbf{\textit{SD}}} & \multicolumn{1}{c}{\textbf{\textit{M}}} & \multicolumn{1}{c}{\textbf{\textit{SD}}} & \multicolumn{1}{c}{\boldmath$\chi^2$} & \multicolumn{1}{c}{\textbf{\textit{p}}} \\
        \midrule
        \textbf{Total} & 18.214 &  7.638 &  11.467 &  4.719 &   7.286  &    0.007{$^{**}$} \\
        \textbf{Task 1} & 3.571 & 0.646 & 2.467 & 1.125 & 9.264  & 0.002{$^{**}$} \\
        \textbf{Task 2} & 2.500 & 1.092 & 2.200 & 0.676 & 0.822  & 0.365\\
        \textbf{Task 3} & 3.571 & 0.646 & 2.133 & 0.990 & 16.084  & $<$0.001{$^{**}$} \\
        \textbf{Task 4} & 2.143 & 1.027 & 1.400 & 1.121 & 3.503  & 0.061{$^{*}$} \\
        \textbf{Task 5} & 1.714 & 1.858 & 1.000 & 1.254 & 1.513  & 0.219 \\
        \textbf{Task 6/1} & 3.214 & 3.093 & 1.400 & 1.639 & 3.740 & 0.053{$^{*}$} \\
        \textbf{Task 6/2} & 1.500 & 2.378 & 0.867 & 1.457 & 0.778  & 0.378  \\
        \bottomrule
        \end{tabularx}
        \vspace{0.1cm}
        \  \small$^{*}$\textit{p} $<$ 0.1 \ \ \ $^{**}$\textit{p} $<$ 0.05
    \end{threeparttable}
\end{table}

The experimental group with AI support scored higher overall than the control group without AI support. On average, the experimental group exhibited a score that was more than six and a half points higher than the control group. Looking at the individual tasks, it can be seen that the AI support did not add significant value in all cases. While the experimental group outperformed the control group in Tasks 1 and 3, this was less significant in Tasks 4 and 6/1. No significant differences were found for the other tasks. Analysis performance did not correlate significantly with participant age ($\chi^2$=1.823; p=0.177) or gender ($\chi^2$=1.910; p=0.167).

In the second part of the analysis task, participants were requested to indicate the probability and level of confidence associated with a given thesis. The assessment of these items differs from that of the first part of the analysis task in that a deviation from the expert base is possible in both directions, i.e., an over- or underestimation of probability and confidence. Consequently, the discrepancy between each individual rating and the mean of the expert ratings for that item was initially determined. The mean (M) and standard deviation (SD) of this discrepancy from the expert judgment were then calculated.

\begin{figure}[ht]
\begin{tikzpicture}
        \pgfplotsset{scale only axis}
        \begin{axis}[
            xmin=1, xmax=19,
            ymin=0, ymax=3,
            xlabel=Item,
            ylabel={Difference to expert judgement},
            width={.86\textwidth},
            height={0.5\textwidth},
            grid=major,
            xtick={1,2,...,19},
            every axis plot/.append style={thick}
            ]
            \addplot[smooth,mark=*]
                coordinates{
                (1,0.842)(2,0.422)(3,0.400)(4,0.983)(5,0.947)(6,0.623)(7,0.467)(8,0.908)(9,1.388)(10,0.407)(11,0.739)(12,0.769)(13,0.301)(14,2.160)(15,1.508)(16,0.593)(17,0.415)(18,0.642)(19,1.652)
                };
            \addplot[smooth,mark=square,blue]
                coordinates{
                (1,0.722)(2,0.873)(3,0.428)(4,2.186)(5,0.823)(6,0.896)(7,0.477)(8,1.077)(9,0.993)(10,0.883)(11,1.174)(12,1.260)(13,0.448)(14,2.343)(15,1.710)(16,0.694)(17,0.517)(18,0.539)(19,1.698)
                }; 
            \addlegendimage{/pgfplots/refstyle=plot_Drang}\addlegendentry{\small Experimental group},
            \addlegendimage{/pgfplots/refstyle=plot_Bewegung}\addlegendentry{\small Control group},
        \end{axis}
        
    \end{tikzpicture}
\caption{Absolute value of the difference between the experimental group and the control group from the probability expert judgment, measured as the standard deviation of the respective item.} \label{Abb7}
\end{figure}
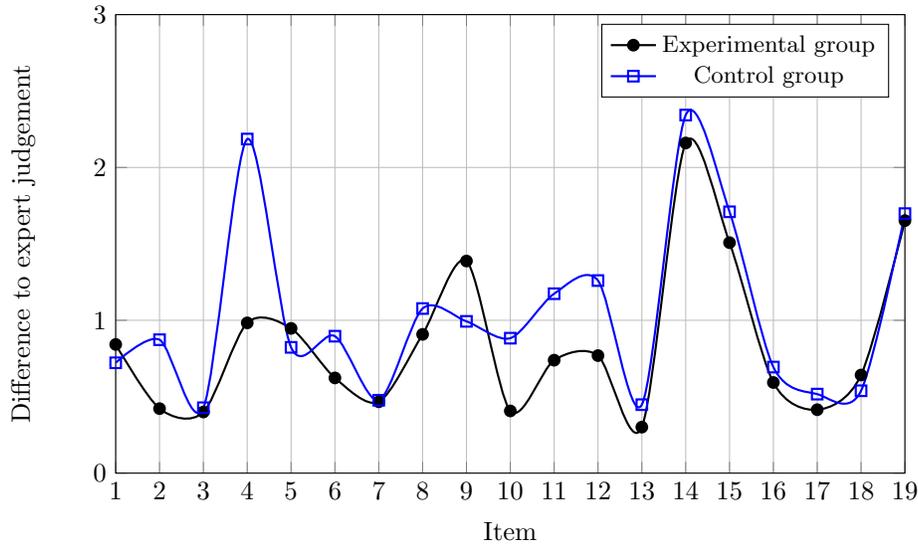

Figure~\ref{Abb7} illustrates the results for the probability ratings, with lower values representing a more favorable outcome.

Although the experimental group still outperformed the control group on most items, the difference is graphically less clear than in the first part of the analysis task. The poorer performance of the control group on Item 4 is still striking, as this was an assessment of whether the agent released at Khan Shaykhun could have been a gas other than sarin, chlorine or a mixture of the two. The experts were almost unanimous in their assessment that this was 'highly unlikely'. 

This view was shared mainly by the experimental group, but not by the control group. The latter rated this thesis on average as 'unlikely'. A similar picture emerges for Items 14 and 15, but here for both groups. In conclusion it can be said that, in contrast to the first part of the analysis task, there is no decline in performance towards the end of the task. 
Probability estimation does not correlate significantly with age ($\chi^2$=3.564; p=0.060) or gender ($\chi^2$=0.688; p=0.407).

Table~\ref{Tab3} shows the results of the two-sided significance tests for independent samples for each task block.
Overall, the experimental group with AI support was able to make significantly more accurate estimates (probability) than the control group. There was no significant difference between the two groups in terms of confidence in their own analysis performance. As with the evaluation of the first part of the analysis task, both probability and confidence were analyzed on a task-by-task basis. It was found that there were no significant differences between the experimental and control groups (therefore, these are not reported in Table~\ref{Tab3}). On a task-by-task basis, only in Tasks 1 and 4 were the AI-assisted participants able to make a significantly more accurate assessment of probability than the control group.

\begin{table}[ht]
\caption{Deviation of the participant's probability and confidence from the expert's judgment, measured as the standard deviation of the item.\bigskip} \label{Tab3}
\begin{threeparttable}
    \renewcommand{\arraystretch}{1.2}
    \setlength{\tabcolsep}{0.15cm}
        \begin{tabularx}{\textwidth}{X rrrrr l }
        \toprule
        & \multicolumn{2}{c}{\textbf{EG}} & \multicolumn{2}{c}{\textbf{KG}} & &  \\
        \cmidrule(rl){2-3} \cmidrule(lr){4-5}
        & \multicolumn{1}{c}{\textbf{\textit{M}}} & \multicolumn{1}{c}{\textbf{\textit{SD}}} & \multicolumn{1}{c}{\textbf{\textit{M}}} & \multicolumn{1}{c}{\textbf{\textit{SD}}} & \multicolumn{1}{c}{\boldmath$\chi^2$}  & \multicolumn{1}{c}{\textbf{\textit{p}}} \\
        \midrule
        \textbf{Probability} & 0.851 & 0.211 & 1.039 & 0.296 & 3.919  & 0.047{$^{**}$} \\
        \textbf{Probability Task 1} & 0.662 & 0.166 & 1.052 & 0.587 & 5.553  & 0.018{$^{**}$} \\
        \textbf{Probability Task 2} & 0.785 & 0.295 & 0.859 & 0.480 & 0.276  & 0.599 \\
        \textbf{Probability Task 3} & 0.921 & 0.310 & 0.849 & 0.324 & 0.399  & 0.528 \\
        \textbf{Probability Task 4} & 0.638 & 0.243 & 1.106 & 0.442 & 10.545  & 0.001{$^{**}$} \\
        \textbf{Probability Task 5} & 1.141 & 0.707 & 1.299 & 0.761 & 0.360  & 0.548 \\
        \textbf{Probability Task 6} & 0.903 & 0.444 & 0.918 & 0.445 & 0.008  & 0.925 \\
        \textbf{Confidence} & 1.091 & 0.248 & 1.039 & 0.356 & 0.220  & 0.639 \\
        \bottomrule
        \end{tabularx}
        \vspace{0.1cm}
        \ $^{**}$\textit{p} $<$ 0.05
    \end{threeparttable}
\end{table}

A post-hoc survey was also conducted to determine how the 14 participants in the experimental group rated the System Usability Scale (SUS) \cite{lewis2018system}, whether the software leads to an improvement in the speed of military analysis, and whether the three AI functions experimentally investigated (AI search, automated summarization, and NER) were considered suitable for use in military analysis.

Table~\ref{Tab4} shows that the participants rated the individual AI functions rather positively in terms of automated summarization (M=4.2), while the AI search (M=3.4) and NER (M=3.1) had more of an average benefit for military analysis. This was measured on a five-point Likert scale. In addition, an increase in analysis speed (M=3.9) was assessed using the AI functions. 

The System Usability Scale (SUS) was rated above average with a score of 86. Participants feel that the user interface can support military analysis efficiently, effectively and satisfactorily. This is in line with the findings by Perboli \textit{et al.} \cite{perboli2021natural}. 

In the context of the labeling of aviation accident documents, the use of AI functions for the partial automation of expert work has reduced the overall investigation time by 30\%.

\begin{table}[t]
\caption{Evaluation of the System Usability Scale (SUS), speed increase, and AI functions by experimental group participants.\bigskip} \label{Tab4}
\centering\begin{threeparttable}
    \renewcommand{\arraystretch}{1.2}
    \setlength{\tabcolsep}{0.15cm}
        \begin{tabularx}{.75\textwidth}{X rr}
        \toprule
        & \multicolumn{1}{c}{\textbf{\textit{M}}} & \multicolumn{1}{c}{\textbf{\textit{SD}}}  \\
        \midrule
        \textbf{AI search} & 3.393 & 0.626  \\
        \textbf{Named Entity Recognition} & 3.143 & 1.420    \\
        \textbf{Automatic summary} & 4.214 & 0.825  \\
        \textbf{Speed} & 3.946 & 1.093  \\
        \textbf{SUS-Score} & 86.250 & 10.504  \\
        \bottomrule
        \end{tabularx}
        \vspace{0.1cm}
    \end{threeparttable}
\end{table}

\section{Discussion}\label{sec6}
\subsection{Comparison of the analytical performance of the experimental group with that of the control group}
\begin{figure}[ht]
\centering\includegraphics[width=0.95\textwidth]{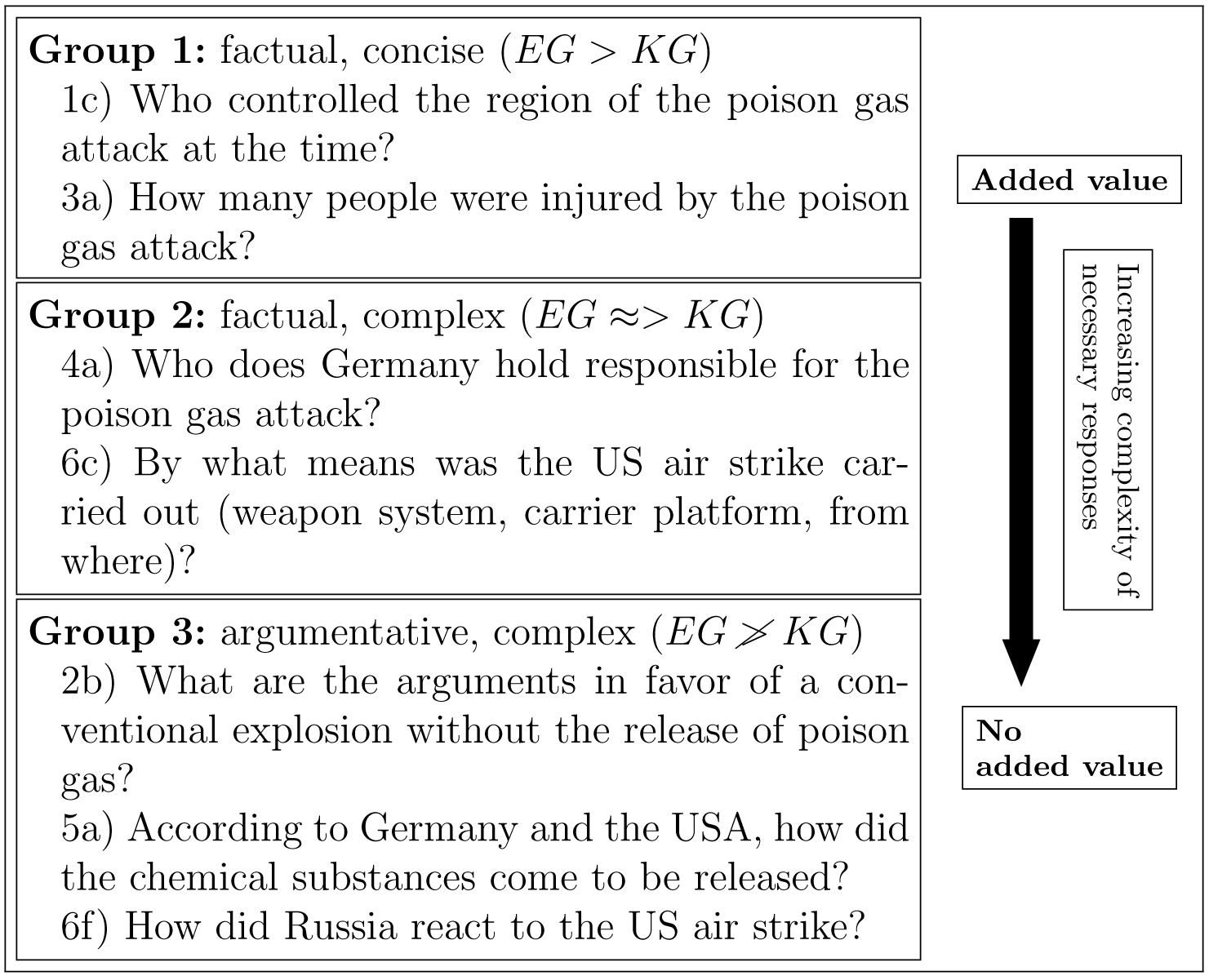}
\caption{Examples of items for each of the three groups of items, in order of complexity of response required.} \label{Abb8}
\end{figure}

For the first part of the analysis task, the experiment showed that the use of AI functions leads to a demonstrable increase in performance. However, a more detailed analysis shows that this is not the case for all tasks. Since these blocks of tasks belong together thematically, further conclusions can be drawn for the AI functions. For this purpose, the tasks are divided up into three groups: Tasks in which the experimental group performed highly significantly (Group 1: Tasks 1 and 3), weakly significantly (Group 2: Tasks 4 and 6/1) and not significantly better (Group 3: Tasks 2, 5 and 6/2). All the questions in Group 1 have in common that they can be answered in a few words and in a factual manner. The sample items shown in Figure~\ref{Abb8} can be answered with 'insurgents' (1c) and 'between 100 and 400' (3a).

Group 2 consists of questions with similar characteristics to Group 1, except that the answers are more complex: Possible answers for Item 4a are 'Syrian government', 'President Assad', or 'Syrian air force', and for Item 6c 'cruise missiles', 'tomahawks', 'warships', and 'destroyers'. All items in Group 2 can be answered largely factually. The items in Group 3, where no statement can be made about the added value of AI functions, require more complex answers. These are mainly argumentative and less clear than those in the first two Groups 1 and 2, requiring comparatively long answers. It can be concluded that the AI functions analyzed provide added value mainly for questions that require a direct and factual answer. As the complexity and ambiguity of the information increases, the benefit decreases.

A similar statement cannot be made for the second part of the analysis task regarding probability. It is true that the participants with AI support were able to give more accurate ratings than the control group on the overall scale and on Tasks 1 and 4. However, this observation is difficult to interpret because the tasks are grouped thematically in the same way, but do not differ in terms of complexity or other characteristics. 

The better performance on probability in Task 4 can be explained by the design of the experiment. This included hypotheses about the US intention behind the air strike on Al Shayrat, about which questions were asked at the end of the first part. 

The poorer analytical performance of the control group, which had already been shown, led to less knowledge about the background of the American air strike. Since the control group had fewer source texts to analyze due to time pressure, their judgments in this regard were less reliable. Nevertheless, the conclusion for Task 4 remains an important observation, as time pressure is a realistic situation for military intelligence. Thus, the overall conclusion that the AI-assisted participants were able to make more accurate judgments than the control group remains valid.

The demonstrably better analysis performance of the AI-assisted experimental group suggests that this is also associated with higher confidence in the analysis performance. However, higher confidence cannot be observed. This can also be seen in the detailed analysis by task block. One explanation for this could be that the confidence assessment in both groups led to uncertainty in the assessment of their own performance due to time pressure. It is possible that the participants had the impression that it was not possible to process all the sources that could have provided further information due to the short time available. This assumption is supported by the fact that participants in both groups rated their confidence significantly lower on average than the experts. The latter had as much time as they needed.

\subsection{Post-hoc survey of the experimental group}
The evaluation of the post-hoc survey shows that the experimental group perceives a speed gain in military analysis through AI functions. This aspect of the post-hoc survey is relevant because, due to the experimental design, speed and workload could not be measured separately but were measured together in the experiment as analysis performance. Theoretically, it could be that the AI functions contribute to a more accurate and comprehensive analysis, but the task itself takes the same amount of time. This would not be a problem, as the improved analysis performance would still be an added value. In this respect, the participants' assessment can be seen as an indication that the speed of intelligence analysis can be increased. This correlation was also confirmed in the personal feedback discussions with the participants. In particular, they felt that the direct answering of questions by the AI search had the potential to increase speed. Comparing the average processing time of 3 hours and 49 minutes for the experts surveyed with the experimental group who had 30 minutes to answer the questions themselves, it can be assumed that the use of AI resulted in a massive time saving. The results of the experimental group were largely identical to those of the experts, especially for the factual questions.

In general, participants rated the automated summary in particular as being above average for use in military analysis. The NER and the AI search, on the other hand, were rated as being of average suitability. Three reasons for this can be deduced from the personal feedback interviews, in which the NER was also criticized by the trial participants: Firstly, the participants could not see any added value in the NER beyond the automatic summarization. Both AI functions were used with the aim of identifying suitable source texts more quickly. However, in a head-to-head comparison, the participants in the experimental group favored the automatic summarization. It can therefore be assumed that the NER is not completely unsuitable for use in AI analysis but appears to be less suitable than the automated summary in a direct comparison due to the overlapping scope of application. Another reason given for the average rating was the sometimes incorrect labeling. In this context, three participants noted that they had wanted to use the NER to identify relevant source texts at the beginning of the study, but then refrained from doing so because, for example, people were incorrectly classified as places. Finally, several participants in the experimental group reported that some of the sub-functions of the NER, such as the color coding in the text or the display on a world map, were perceived as useful, but then simply overlooked due to the number of other features. Participants also attributed this to the time constraints of the experiment. In summary, the majority of respondents rated the NER as having average added value to the military analysis process due to competition from automated summarization, functionality that still needs improvement in some cases, and other AI functions.

The System Usability Scale (SUS) was rated above average by the participants. This shows that the user interface of the deepCOM demonstrator was successfully designed to be intuitive. This is also confirmed by the experience of getting used to deepCOM before the experiment. Not only did this rarely take longer than five minutes, but there were generally no questions about the user interface as it was perceived as self-explanatory. In the personal feedback interviews at the end of the experiment, the participants were also unable to make any suggestions for improving the design of the user interface.

\subsection{The challenges associated with the use of artificial intelligence in military analysis}
The study identified several challenges in applying AI to the military analysis process. For example, the AI search encountered problems when the underlying texts contained no or insufficient information for the answer. LLM always gives the answer with the highest probability. If the sources are poor, this can sometimes lead to an accumulation of wrong answers. This was also evident in the first part of the analysis task, which contained two trick questions. The correct answer to the question 'Which nations were also involved in the US air strike?' was 'None', and the possible answer to the question 'What are the signs of a conventional explosion without the release of poison gas?' was 'Nothing'. In both cases, the AI search gave the wrong answer. The problem is that LLM always finds a passage in the source text that could answer the question posed.

There is also room for improvement in the NER. Although entities are almost always recognized and extracted as such, the classification into different groups (person, place, etc.) still fails too often. Lemmatization is not perfect in some cases, either, especially in the case of military terms and proper names, which are reduced to incorrect stems. These problems could probably be solved by retraining. The NER used was not adapted for a specific purpose but was trained for German texts in general. However, it could be specialized for entities that occur exclusively in military source texts (e.g. names of weapons, names of military leaders). This would require retraining on as large a dataset of military reports as possible. The automatic creation of domain-specific dictionaries based on military reports would be another possible improvement in this context \cite{haffner2023introducing}.

A final challenge is the volume of text to be processed in military intelligence. The analysis scenario of the experiment involved only 50 reports, a quantity that can also be processed with a BOW search in a reasonable amount of time. Scaling up to several thousand texts, for example, could quickly overwhelm the BOW search, which is based on word frequency and does not really capture the content of the texts. The semantic understanding of an LLM, on the other hand, allows the same word occurrences to be interpreted in the context of the text or paragraph, making larger volumes of text manageable. In the subsequent phase, more sophisticated AI algorithms will undertake extractive summarization across multiple documents, employing unsupervised techniques such as \cite{lamsiyah2021unsupervised}. Utilizing sentence embedding representations, they will identify pivotal sentences based on a combined scoring system. This system will then assess the relevance, novelty, and positional importance of sentence content, ensuring that the summary encompasses the most crucial and diverse information from the documents.  The fact remains, however, that the performance of any analysis-supporting AI can never be better than the quality of the military source reports it works with \cite{vogel2021impact}. So, while the use of AI makes it possible to sift through and analyze more sources, AI does not guarantee that the sources are reliable \cite{horlings2023dealing}. More is not always better.

\subsection{Limitations of the design of the study}
One limitation to the validity of this study is that the time allowed for completion was limited to 30 minutes. The experts were used as a reference for the assessment of analytic performance but were not under time constraints in completing the analysis tasks. In practice, however, it is often the case that intelligence products are produced under great time pressure. In this respect, the restriction of the processing time for the participants in the experiment can be regarded as a real situation in practice. In the first part of the analysis task, the experimental group achieved on average 100\% of the baseline for two items, i.e., they performed as well as the experts. It cannot be excluded that the experimental group performed better than the experts due to the support of the AI functions, but that this could not be measured. It is also likely that the participants' answers to the confidence and probability questions were influenced by the time limit.

Another critical point regarding the study design is that the added value of the AI functions was only demonstrated in relation to a single analysis scenario. The scenario was designed to best model a real-world military intelligence scenario based on unclassified sources. In particular, future studies should critically examine the extent to which the specified time period in which all source texts were available and the nature of the sources influenced the results of the experiment. All texts in the analysis scenario were published within four days, between April 4 (the Khan Shaykhun attack) and April 7, 2017 (notification of the need for information by military decision-makers). As part of their analysis, the participants were confronted with a partially misleading information situation. They had to work with source texts that also contained uncertain and contradictory information. For example, it is unclear what an extension of the time corridor from four days to several years means for the added value of AI functions. Even if distractors were taken into account, the added value of AI functions would have to be considered in the light of source texts that are of general military relevance, but not for the analysis just conducted.

Finally, it should be noted that due to the design of the study, it is not possible to make statements about the AI functions individually, as their added value was only analyzed in combination. It is possible that one of the AI functions did not add value on its own, but only benefited from the added value of the others. To account for this, only three AI functions were included in the experimental study from the outset. Conceptually, these three functions provided added value to the military analysis. An additional post-hoc survey was conducted to take this into account. However, these are subjective assessments of the participants and do not allow any causal conclusions to be drawn. Therefore, a limitation of the experiment that cannot be eliminated without further research is that while the three AI functions as a whole provide significant added value, the added value of the individual functions is not necessarily given.

\section{Conclusion}\label{sec7}
A demonstrator was used to test three AI functions designed to support the work of military analysts in providing the most accurate situational awareness possible. These AI functions are essentially AI functions that have been made possible by advances in the field of LLM. In addition, there is a wide range of other possible applications of AI in military analysis that could not be included in this study in the interest of brevity. A further increase in the performance of the analyzed AI functions can be expected from new developments, especially in the training of German or European LLM. Data protection and confidentiality also play a central role in the use of artificial intelligence in military analysis. The present study has shown experimentally that the three AI functions of AI search, NER, and automated summarization in combination provide added value for military analysts. Not only does analysis performance increase, but so does the ability to make more accurate assessments. In particular, the participants in the experiment saw an advantage in the increased speed of military analysis.

Using the intelligence cycle, the AI functions were positioned within military intelligence analysis. It was found that there are numerous potential areas of application for the AI functions proposed in this thesis alone. In practice, however, it is often necessary to specify which AI functions can provide concrete support in military analysis \cite{cho2020priority}. However, there are also areas of military analysis that may never be supported by AI due to ethical considerations \cite{blanchard2023ethics}. Future research could complement the present study in that the added value of AI in supporting analysis also applies to the application of other AI functions. Extending the findings to different analysis scenarios could also contribute to the generalizability of the results. In the future, the necessity for trust and transparency in AI systems, particularly in the context of military applications, highlights the requirement for well-defined methodologies to address the ``black box" nature of generative AI. The provision of such clarity will facilitate a more nuanced understanding and trust in AI decisions and the ``rational" actions they generate among stakeholders, including customers, developers, and regulators. While not a prerequisite in the transparent study of well-understood documents, the application of Explainable AI (XAI) becomes imperative in more intricate, real-world scenarios. XAI enhances the interpretability and user-friendliness of AI systems, fosters trust in their decision-making processes, and aligns development with ethical and regulatory standards \cite{chamola2023review,dwivedi2023explainable}.

\appendix

\section{Sources of the texts for the analysis scenario}\label{annex1}
\begin{flushleft}
\begin{itemize}
\item Suspected poison gas attack: US and EU accusations against the Assad government. (2017). Heise Online. Retrieved from https://www.heise.de/tp/features/Mutmasslicher-Giftgasangriff-Vorwuerfe-der-USA-und-der-EU-gegen-die-Regierung-Assad-3675216.html?seite=all
\medskip

\item One lie and many inconsistencies: Russian statement on the poison gas attack in Syria. (2017). Spiegel Online. Retrieved from https://www.spiegel.de/politik/ausland/giftgasangriff-in-syrien-eine-luege-und-viele-ungereimtheiten-a-1141982.html
\medskip

\item Syria: ``This chemical attack could only cause a massacre". (2017). Libération. Retrieved from https://www.liberation.fr/planete/2017/04/05/syrie-cette-attaque-chimique-ne-pouvait-que-provoquer-un-massacre\_1560802/
\medskip

\item Deadly chemical attack in Syria: Why everything blames Damascus. (2017). Le Parisien. Retrieved from https://www.leparisien.fr/international/attaque-chimique-meurtriere-en-syrie-pourquoi-tout-accuse-damas-05-04-2017-6826935.php
\medskip

\item Ministry of Defense of the Russian Federation: Syrian aviation bombed a chemical weapons production workshop in Khan Sheikhun. (2017). TASS. Retrieved from https://tass.ru/politika/4154524
\medskip

\item Experts: Allegations of the use of chemical weapons by Damascus are not confirmed by facts. (2017). RIA Novosti. Retrieved from https://ria.ru/20170405/1491556253.html
\medskip

\item Syrian aircraft attacked a chemical weapons depot in Idlib. (2017). RIA Novosti. Retrieved from https://ria.ru/20170405/1491508627.html
\medskip

\item Worst chemical attack in years in Syria; US blames Assad. (2017). The New York Times. Retrieved from https://www.nytimes.com/2017/04/04/world/middleeast/syria-gas-attack.html?searchResultPosition=1
\medskip

\item White House condemns ``heinous" attack in Syria. (2017). Fox News. Retrieved from https://www.foxnews.com/politics/white-house-condemns-heinous-attack-in-syria
\medskip

\item Trump says chemical attack in Syria crossed many lines. (2017). Reuters. Retrieved from https://www.reuters.com/article/us-mideast-crisis-syria-idlib-idUSKBN1770YU
\medskip

\item Iran condemns the use of chemical weapons in Syria. (2017). Islamic Republic News Agency. Retrieved from https://en.irna.ir/news/82483290/Iran-condemns-use-of-chemical-weapons-in-Syria
\medskip

\item USA blames Russia and Iran for use of poison gas. (2017). Tagesspiegel. Retrieved from https :// www .tagesspiegel.de/politik/giftgasangriff-in-syrien-usa-geben-russland-und-iran-mitverantwortung-fuer-giftgaseinsatz/19610606.html
\medskip

\item US air strike against airbase in Syria. (2017). Nordkurier. Retrieved from https://www.nordkurier.de/nachrichten/ticker/us-luftschlag-gegen-luftwaffenstuetzpunkt-in-syrien-0727523704.html
\medskip

\item Syrian war: Global reaction to US missile attack. (2017). British Broadcasting Corporation. Retrieved from https://www.bbc.com/news/world-us-canada-39526089
\medskip

\item Mapping the targets of the US military attack on Syria. (2017). The New York Times. Retrieved from https://www.nytimes.com/interactive/2017/04/07/world/middleeast/us-syria-missile-attack-airbase.html
\medskip

\item ``First they lied, then they attacked": Syrians on the US strike on the airbase. (2017). RIA Novosti. Retrieved from https://ria.ru/20170407/1491759926.html
\medskip

\item The USA launches cruise missiles at the Syrian airbase. (2017). Al Jazeera. Retrieved from https://www.aljazeera.com/news/2017/4/7/us-launches-cruise-missiles-on-syrian-airbase
\medskip

\item Seven decades, seven facts: US policy toward Syria in a nutshell. (2017). Deutsche Welle. Retrieved from https://www.dw.com/en/seven-decades-seven-facts-us-policy-on-syria-in-brief/a-38346847
\medskip

\item A city rehearses the council system. (2017). WOZ The weekly newspaper. Retrieved from https://www.woz.ch/-79f0
\medskip

\item Refueat: Food trucks were yesterday, now comes the food bike. (2017). Berliner Zeitung. Retrieved from https://www.berliner-zeitung.de/mensch-metropole/refueat-foodtrucks-waren-gestern-jetzt-kommt-das-foodbike-li.67991
\medskip

\item Poison gas has often played a role in the conflict. (2017). Swiss radio and television. Retrieved from https://www.srf.ch/news/international/giftgas-spielte-im-konflikt-schon-oft-eine-rolle
\medskip

\item Murdering people is apparently no longer ok when poison gas is involved. (2017). The Postillion. Retrieved from https://www.der-postillon.com/2017/04/giftgas-syrien.html
\medskip

\item Propaganda on the poison gas attack in Syria: Reality is now just an opinion. (2017). Spigel Online. Retrieved from https://www.spiegel.de/netzwelt/netzpolitik/propaganda-beim-giftgasangriff-in-syrienkonflikt-kolumne-von-sascha-lobo-a-1141980.html
\medskip

\item Assad's history of chemical attacks and other atrocities. (2017). The New York Times. Retrieved from https://www.nytimes.com/2017/04/05/world/middleeast/syria -bashar -al -assad -atrocities -civilian -deaths -gas -attack.html
\medskip

\item Assad tells newspaper he sees no ``option but victory" in Syria. (2017). Reuters. Retrieved from https://www.reuters.com/article/us-mideast-crisis-syria-assad-idUSKBN1780W7
\medskip

\item The situation in Syria constitutes an international armed conflict - Red Cross. (2017). Reuters. Retrieved from https://www.reuters.com/article/us-mideast-crisis-syria-redcross-idUSKBN17924T
\medskip

\item US strike on Syria means attacking powers against terrorism: Assad adviser. (2017). Islamic Republic News Agency. Retrieved from https://en.irna.ir/news/82485257/US-strike-on-Syria-means-attacking -powers-fighting-terrorism
\medskip

\item The United Nations called for a 72-hour humanitarian ceasefire in Syria. (2017). RIA Novosti. Retrieved from https://ria.ru/20170406/1491636653.html
\medskip

\item Russia will train local pioneers in Homs. (2017). RIA Novost. Retrieved from https://ria.ru/20170407/1491698378.html
\medskip

\item MARKET: WALL STREET STARTS WITH CAUTION. (2017). TradingsSat. Retrieved from https://www.tradingsat.com/actualites/marches-financiers/marche-wall-street-debute-sur-une-note-de-prudence-732983.html
\medskip

\item Trump ignores human rights, they remember him. (2017). Nouvel Obs. Retrieved from https://www.nouvelobs.com/monde/l-amerique-selon-trump/20170406.OBS7637/trump-ignore-les-droits-de-l-homme-ils-se-rappellent-a-lui.html
\medskip

\item Syria chemical weapons timeline: A history of the Assad regime attacking civilians. (2017). International Business Times. Retrieved from https://www.ibtimes.com/syria-chemical-weapons-timeline-history-assad-regime-attacking-civilians-2522011
\medskip

\item What is Bashar al-Assad thinking (2017). The New York Times. Retrieved from https://www.nytimes.com/2017/04/07/opinion/what-is-bashar-al-assad-thinking.html
\medskip

\item US strike on Syria changes North Korea dynamic as Trump, Xi, meet. (2017). The Japan Times. Retrieved from https://www.japantimes.co.jp/news/2017/04/07/world/u-s-strike-syria-changes-north-korea-dynamic-trump-xi-meet/
\medskip

\item Israeli leaders praise wall-to-wall US strike on Syria. (2017). The Times of Israel. Retrieved from https://www.timesofisrael.com/netanyahu-praises-message-of-resolve-in-us-strike-on-syria/
\medskip

\item IRC responds to Idlib attacks: How can you help? (2017). International Rescue Committee. Retrieved from https://www.rescue.org/article/irc-responds-idlib-attack-how-you-can-help
\medskip

\item At least 9 chemical attacks since the beginning of 2017 (2017). Syrian Network for Human Rights. Retrieved from https://sn4hr.org/arabic/2017/04/05/7860/
\medskip

\item Zarif criticizes US hypocrisy in standing up for terrorists. (2017). More News Agency. Retrieved from https://en.mehrnews.com/news/124620/Zarif-slams-US-hypocrisy-in-siding-with-terrorists
\medskip

\item What has changed? The role of photography in the Syrian war. (2017). Retrieved from https://time.com/4729560/syria-chemical-attack-photography-impact/
\medskip

\item Syrians forget their emigration plans in Egypt. (2017). TIME. Retrieved from https://fr.africanews.com/2017/04/07/en-egypte-des-syriens-oublient-leurs-projets-d-emigration/
\medskip

\item Women of Jihad: ``I was with a Kalashnikov", recalls the ex-recruiter. (2017). Le Parisien. Retrieved from https://www.leparisien.fr/faits-divers/femmes-du-djihad-je-sortais-avec-un-kalachnikov-se-souvient-l-ex-recruteuse-06-04-2017-6828954.php
\medskip

\item UN warns of ``desperate situation" for Syrian refugees. (2017). L'Orient-Le Jour. Retrieved from https://www.lorientlejour.com/article/1044667/lonu-alerte-sur-la-situation-desesperee-des-refugies-syriens.html
\medskip

\item Wall Street is disappointed by the new Fed member's comments. (2017). Investor. Retrieved from https://www.investor.bg/sasht/337/a/wall-street-se-razocharova-ot-komentarite-na-noviia-chlen-na-fed-237251/
\medskip

\item Shooting in Baghdad: Why 7 years later the perpetrators of the war crime have not been named in the U.S. (2017). Russia Today. Retrieved from https://russian.rt.com/world/article/375637-wikileaks-ssha-voennye-prestupleniya-vina
\medskip

\item The dark history of chemical weapons. (2017). Cicero. Retrieved from https://www.cicero.de/aussenpolitik/giftgaseinsatz-in-syrien-die-duestere-geschichte-der-chemiewaffen
\medskip

\item Actors in Syria: Proxy war for many powers. (2017). Tagesschau. Retrieved from https://www.tagesschau.de/ausland/akteure-syrien-101.html
\medskip

\item Oil fluctuations after US missile attack in Syria. (2017). Upstream. Retrieved from https://www.upstreamonline.com/online/oil-surges-after-us-missile-strike-in-syria/2-1-66043

\end{itemize}
\end{flushleft}

\section{Complete list of Items}\label{annex2}
\subsection{Questions about the first part of the analysis task}
\begin{enumerate}
\item According to media reports, a poison gas attack has taken place in Syria.
\begin{compactenum}[(a)]
\item Where did the poison gas attack take place? (city, governorate)
\item When did the poison gas attack take place? (date, approximate time/time of day)
\item Who controlled the region of the poison gas attack at the time?
\medskip
\end{compactenum}
\item Release of poison gas.
\begin{compactenum}[(a)]
\item What speaks for the release of poison gas?
\item What are the arguments in favor of a conventional explosion without the release of poison gas?
\item Which chemical substances are said to have been used in the attack?
\medskip
\end{compactenum}
\item Effect of the attack.
\begin{compactenum}[(a)]
\item How many people were injured by the poison gas attack? (approximate range from-to) 
\item How many people were killed by the poison gas attack? (approximate range from-to)
\item What was the status of the injured and dead under international law? (e.g., civilian, insurgent, combatant)
\item What material damage was caused by the poison gas attack?
\medskip
\end{compactenum}
\item Perpetrator of the attack.
\begin{compactenum}[(a)]
\item Who does Germany hold responsible for the poison gas attack?
\item Who does the USA hold responsible for the poison gas attack?
\item Who does Russia hold responsible for the poison gas attack?
\medskip
\end{compactenum}
\item Course of the release of the chemical substances.
\begin{compactenum}[(a)]
\item According to Germany and the USA, how did the chemical substances come to be released?
\item According to Russia and Syria under President Assad, how did the chemical substances come to be released?
\medskip
\end{compactenum}
\item The USA responded to the poison gas attack with an air strike.
\begin{compactenum}[(a)]
\item What was the target of the US air strike? (location, facility, under whose control)
\item Which nations also took part in the US air strike?
\item By what means was the US air strike carried out (weapon system, carrier platform, from where)? 
\item How did Germany react to the US air strike?
\item How did Syria under President Assad react to the US air strike? 
\item How did Russia react to the US air strike?
\end{compactenum}
\end{enumerate}

\subsection{Questions about the second part of the analysis task}

\begin{enumerate}
\item Released chemical gas.

\begin{compactenum}[(a)]
\item The chemical gas released is exclusively sarin.
\item The chemical gas released is exclusively chlorine.
\item The chemical gas released is a mixture of sarin and chlorine.
\item The chemical gas released is a different, unnamed chemical gas.
\medskip
\end{compactenum}

\item Course of the release of the poison gas.

\begin{compactenum}[(a)]
\item The course of events as described by the USA/Germany (poison gas attack by fighter planes of the Syrian air force under Assad) is correct.
\item The course of events as described by Russia/Syria (accidental bombing of a rebel poison gas storage site/factory by the Syrian air force using conventional weapons) is correct.
\medskip
\end{compactenum}

\item Damage caused by the US air strike on the Syrian military airfield Al Shay\-rat.

\begin{compactenum}[(a)]
\item After the air strike, the airfield can no longer be used for take-offs and landings until it has been repaired.
\item Chemical warfare agents were destroyed in the air strike.
\item Syrian Air Force fighter jets were destroyed in the air strike.
\medskip
\end{compactenum}

\item 
Intention of the USA behind the air strike.
\begin{compactenum}[(a)]
    \item 
    The air strike was intended to signal the general military strength and readiness of the USA to intervene ("show of force").
    \item 
    The air strike was intended as a concrete warning to the Assad government that (possible further) chemical attacks would be met with military force.
    \item 
    The air strike was intended to reduce the Assad government's ability to carry out (possible further) chemical attacks.
\medskip
\end{compactenum}

\item Possible future reactions of Syria under President Assad to the US air strike.

\begin{compactenum}[(a)]
\item Syria will demonstrate its independence and freedom of action to the international community through a (possible further) air strike with chemical warfare agents.
\item Under international pressure, Syria will admit (co-)responsibility for the attack in Khan Shaykhun.
\item Syria insists on its account of the accidental bombing of a rebel poison gas storage or production site. Syria will therefore intensify its military efforts against the latter.
\item In view of the increasing implausibility of its account of how the attack in Khan Shaykhun took place, Syria will blame others than the rebels but will not admit any guilt itself.
\medskip
\end{compactenum}

\item Possible future Russian reactions to the US air strike.

\begin{compactenum}[(a)]
\item Russia will continue to expand its air defense capabilities in Syria.
\item Russia will not take any further steps until the United Nations investigation into the attack has been completed.
\item Russia will respond with a military counterattack against US allies in the region.
\end{compactenum}

\end{enumerate}

%
%
%
%

\end{document}